\documentclass{article}

\usepackage[nonatbib, final]{neurips_2019}

\usepackage[utf8]{inputenc} 
\usepackage[T1]{fontenc}    
\usepackage{hyperref}       
\usepackage{url}            
\usepackage{booktabs}       
\usepackage{amsfonts}       
\usepackage{nicefrac}       
\usepackage{microtype}      

\usepackage{amsmath}
\usepackage{amssymb}
\usepackage{algorithmic}
\usepackage{algorithm}
\usepackage{graphicx}
\usepackage{textcomp}
\usepackage{subcaption}
\usepackage{xcolor}
\usepackage{hyperref}
\usepackage[numbers, sort, comma, square]{natbib}
\usepackage{caption}
\usepackage{subcaption}
\usepackage{todonotes}

\title{Detecting Patterns of Physiological Response to Hemodynamic Stress via Unsupervised Deep Learning}

\author{
  Chufan Gao \\
  Carnegie Mellon University\\
  Pittsburgh, PA 15213 \\
  \texttt{chufang@andrew.cmu.edu} \\
  \And
  Fabian Falck \\
  Carnegie Mellon University\\
  Pittsburgh, PA 15213 \\
  \texttt{ffalck@andrew.cmu.edu} \\
  \And
  Mononito Goswami \\
  Carnegie Mellon University\\
  Pittsburgh, PA 15213 \\
  \texttt{mgoswami@andrew.cmu.edu} \\
    \And
  Anthony Wertz \\
  Carnegie Mellon University\\
  Pittsburgh, PA 15213 \\
  \texttt{awertz@cmu.edu} \\
   \And
  Michael R.\ Pinsky \\
  University of Pittsburgh\\
  Pittsburgh, PA 15213 \\
  \texttt{pinsky@pitt.edu} \\ 
  \And
  Artur Dubrawski \\
  Carnegie Mellon University\\
  Pittsburgh, PA 15213 \\
  \texttt{awd@cs.cmu.edu}
}

\begin{document}

\maketitle

\begin{abstract}
Monitoring physiological responses to hemodynamic stress can help in determining appropriate treatment and ensuring good patient outcomes. Physicians' intuition suggests that the human body has a number of physiological response patterns to hemorrhage which escalate as blood loss continues, however the exact etiology and phenotypes of such responses are not well known or understood only at a coarse level. Although previous research has shown that machine learning models can perform well in hemorrhage detection and survival prediction, it is unclear whether machine learning could help to identify and characterize the underlying physiological responses in raw vital sign data. We approach this problem by first transforming the high-dimensional vital sign time series into a tractable, lower-dimensional latent space using a dilated, causal convolutional encoder model trained purely unsupervised. Second, we identify informative clusters in the embeddings. By analyzing the clusters of latent embeddings and visualizing them over time, we hypothesize that the clusters correspond to the physiological response patterns that match physicians' intuition. Furthermore, we attempt to evaluate the latent embeddings using a variety of methods, such as predicting the cluster labels using explainable features.
\end{abstract}

\section{Introduction}
\textit{Internal hemorrhage}, also known as internal bleeding, is defined as leakage of blood from the circulatory system into surrounding tissue and neighboring body cavities \cite{ho2005rapid} and is the ``most frequent complication of a major surgery'' \cite{lozano2012global}. It causes an estimated 1.9 million deaths worldwide annually \cite{lozano2012global, falck2018deephemorrhage}. Survivors of hemorrhage suffer from poor long-term adverse outcomes, such as multiple organ failures and an increased mortality rate \cite{mitra2014long, halmin2016epidemiology}. Thus, analyzing hemorrhage and its underlying physiological patterns could provide clinicians with a better understanding of the progression of a bleed event, inform effective interventions, and help reduce adverse outcomes.

Machine learning can provide a framework to harness the high-dimensional, raw time series data produced by the vital sign monitoring equipment and yield valuable medical insights. For example, \citet{falck2018deephemorrhage} first showed that the prediction of hemorrhage is reliably possible from raw, multivariate vital sign data using recurrent and convolutional neural networks, while being generally inferior compared to a random forest classifier trained on handcrafted statistical features. \citet{nagpal2019hemo} showed that Emission Models can outperform logistic regression, random forest, and a multilayer perceptron in early hemorrhage detection purely based on central venous pressure (CVP). \citet{zambrano2015detection} demonstrated that a random forest classifier trained on features from analyzing shapes of arterial blood pressure waveforms is able to detect bleeding. \citet{li2018graph} predicted hemorrhagic shock in pigs from prebleed blood draw data by analyzing Graphs of Temporal Constraints (GTC)~\cite{guillame2017classification}.

While prior work has focused on rapid detection of hemorrhage and survival prediction, research on understanding the underlying changes in physiological responses associated with blood loss is lacking. \textit{Supervised} learning techniques are not applicable to discover these patterns, since ground truth data on physiological states is unavailable. Using \textit{unsupervised} learning techniques, however, allows us to examine unlabeled physiological data and may enable the discovery of such state changes. \citet{lei2017learning}s' work demonstrated that such interesting patterns could be found from CVP through correlation clustering. These observed clusters, however, correspond to broad physiological states -- one cluster corresponded mostly to the pre-bleed phase, a second cluster would take over after bleeding started, and a third cluster would appear even further into the bleed -- which is why we urge for a more finer-grained grouping, in particular based on multiple vital signs.

\begin{figure}[t!]
    \centering
    \includegraphics[width=\linewidth]{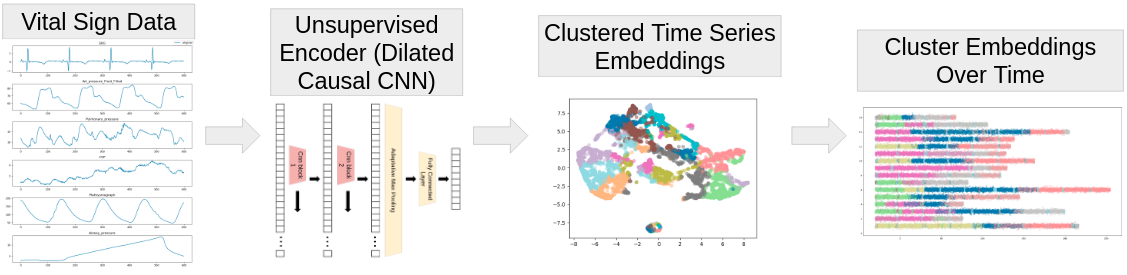}
    \caption{Overview of the methodology. First, we train an encoder model on the normalized raw bleed data to obtain embeddings. Then, we cluster them and visualize data assignments to clusters over time to reveal the sequential structure of physiologic responses to hemodynamic stress.}
    \label{fig:overview}
\end{figure}

In this paper, we propose an embedding algorithm using a state-of-the-art deep unsupervised \textit{dilated, causal} convolutional encoder model \cite{franceschi2019unsupervised} to find informative embeddings from continuous vital sign time series hemorrhage data of six vital signs. We use agglomerative clustering to obtain groups (clusters) of latent embeddings that may correspond to different physiological states that the model detects during the observation period.
A schematic overview of our methodological pipeline is shown in Fig.~\ref{fig:overview}. 
Our contributions are two-fold: (1) An embedding algorithm with a novel sampling methodology (described in more detail in Appendix \ref{app:Sampling methodology}) that encodes high-dimensional, raw vital sign data into lower-dimensional embeddings, and (2) an analysis of the hypothesized, underlying physiological response patterns of subjects through clustering of the embeddings.

\section{Methodology}

\subsection{Data}

Our data consists of vital sign measurements of 16 healthy pigs\footnote{The study protocol was approved by the University of Pittsburgh IACUC.}. We used pigs because their hemorrhage data is more readily available, and expert opinion show that their response to hemorrhage is similar to humans. They are anesthetized and let to rest for a 30 minute period to establish a stable baseline period. Following this baseline period, the pigs are bled at a rate of $5~ \nicefrac{mL}{min}$ until they reach a mean arterial pressure (MAP) of 40 mmHg. The rate of $5~\nicefrac{mL}{min}$ mimics what might be expected in a difficult to detect, occult bleed post-surgery. The time series data also contains irregular laboratory blood draws. We used 6 physiologic measurements (captured at 250 Hz): aortic, pulmonary arterial, and central venous blood pressures (ART, PAP, CVP, respectively), electrocardiogram (ECG), photoplethysmograph from a pulse oximeter, and airway pressure. Physicians' intuition indicate that they may contain important semantic information about the physiological status of the pigs. In addition, blood draws for laboratory testing were performed on each pig regularly throughout the entire experiment. The data collection methodology follows~\citet{pinsky1984instantaneous}.

\subsection{Causal Dilated Convolutional Neural Network}

For our deep unsupervised embedding model, we use a convolutional encoder as proposed by \citet{franceschi2019unsupervised}, which is inspired by \citet{bai2018empirical}'s Temporal Convolutional Networks (TCN). The encoder is able to obtain meaningful embeddings that perform well on time series classification and regression tasks and trains significantly faster than a traditional RNN encoder-decoder model \cite{franceschi2019unsupervised}. We choose to use this model as opposed to traditional statistical feature extraction, since our goal is to automatically discover interest patterns with no assumptions on the underlying hemorrhage data. The generic dilated, causal convolution architecture is depicted in Appendix \ref{app:Additional Methdology}, Fig.~\ref{fig:model}. We train this model in an unsupervised fashion with triplet loss (details provided in Appendix \ref{app:Triplet Loss}).  Our sampling algorithm extracts positive, negative, and reference samples, as shown Appendix \ref{app:Sampling methodology} in Fig.~\ref{fig:sampling} and Algorithm \ref{alg:sampling}. To obtain embeddings of the bleed sequences, we split the time series hemorrhage data of each subject into nonoverlapping time windows of 600 timesteps. This allows us to get an embedding of the vital sign sequence data for every 2.4 seconds. The number of timesteps is a hyperparameter we found that provided reasonable computation times while yielding interesting results. In total, we obtain \textasciitilde 70,000 time windows.  Furthermore, we explore two different training methodologies. (1) We let the model discriminate between subjects by sampling training windows across different subjects (allowing it to consider both intra-subject and inter-subject differences). (2) We restrict sampling of training windows within a subject (allowing it to only learn intra-subject differences).

\subsection{Clustering and Evaluation}
Since we have no ground truth, validation of these embeddings is difficult. However, by clustering the data along the time dimension, we can evaluate the embeddings indirectly by comparing the sequential clusters with clinical rational. We use agglomerative clustering with Ward linkage \cite{mullner2011modern}. Intuitively, adjacent time windows within one subject's data should belong to the same cluster. We also look for consistency in the order of the clusters over time across subjects. In an attempt to explain these clusters, we also use a random forest classifier and a 2 layer fully connected network to predict cluster labels using explainable features -- mean, median, standard deviation, range, and power transforms of the vital signs. Finally, we also show robustness of our encoder by 4-fold cross-validation (by subject) in Figure \ref{fig:clusters_cv}.

\section{Results}
\begin{figure*}[h!]
\centering
\begin{subfigure}[h]{0.495\textwidth}
    \label{fig:a}
    \includegraphics[width=\linewidth]{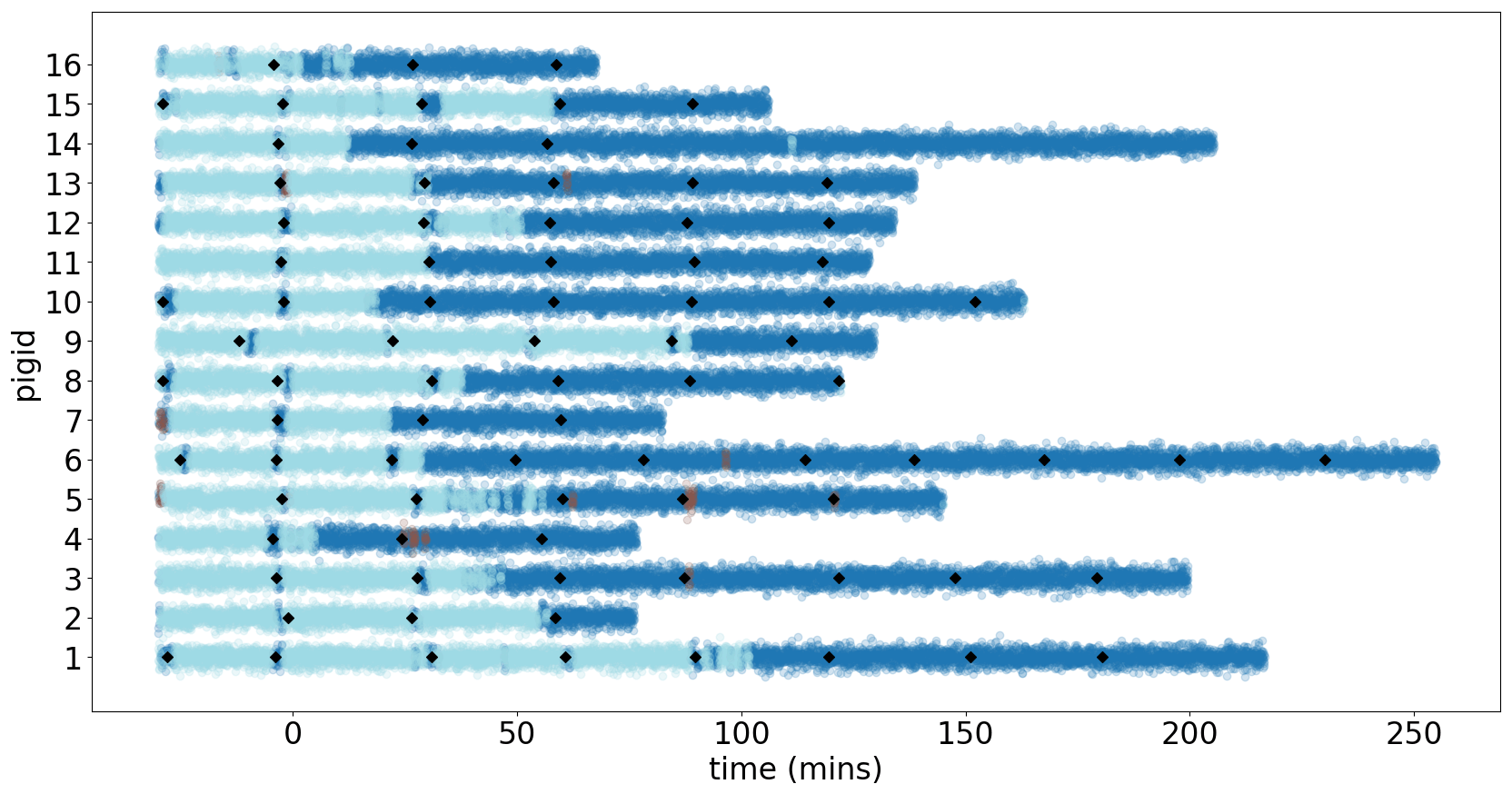}
    \caption{3 clusters,  considering intra-subject and inter-subject differences}
    \label{fig:b}
    \includegraphics[width=\linewidth]{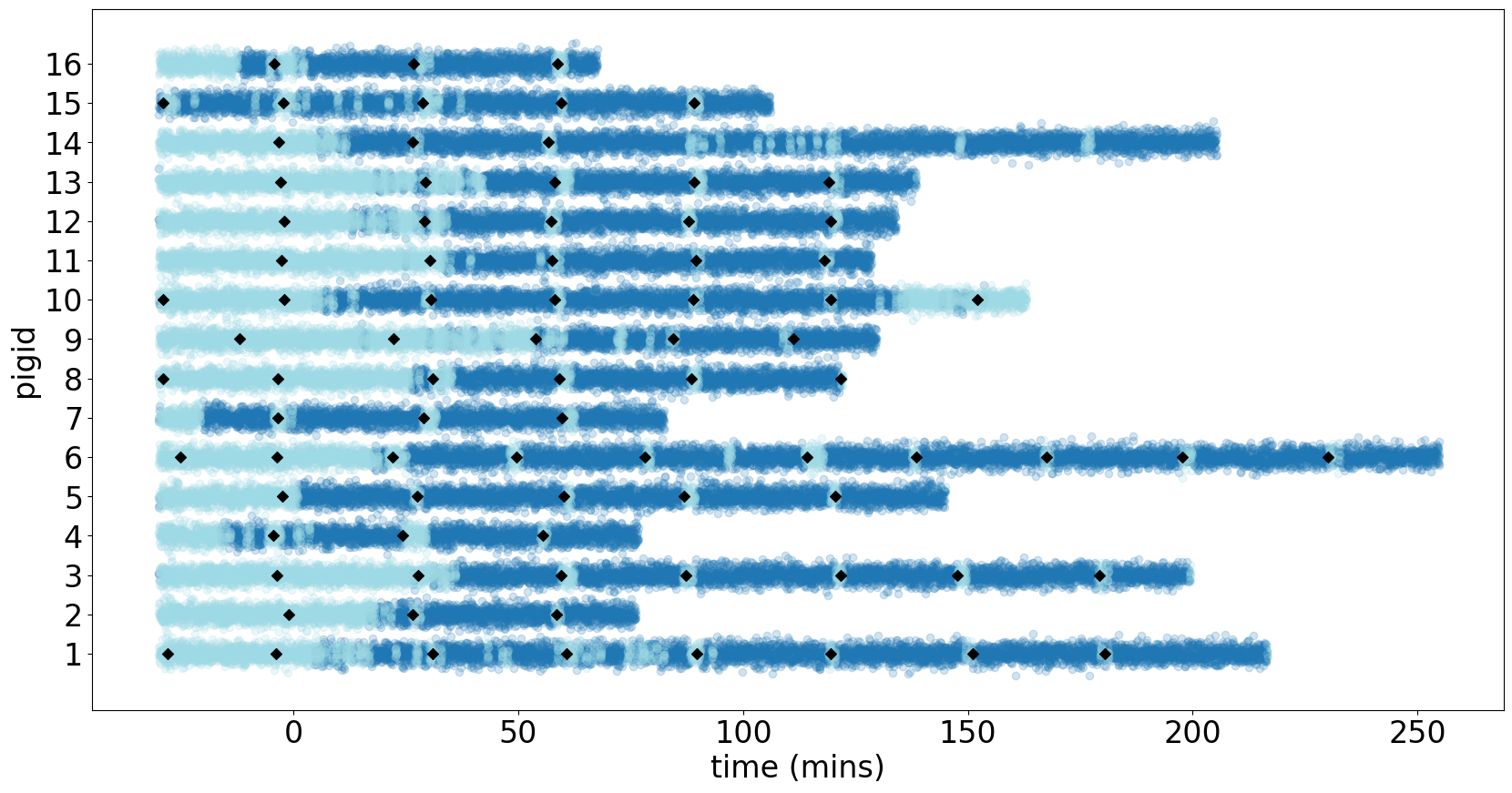}
    \caption{2 clusters, considering intra-subject differences only}
\end{subfigure}%
\hfill
\begin{subfigure}[h]{0.495\textwidth}
    \includegraphics[width=\linewidth]{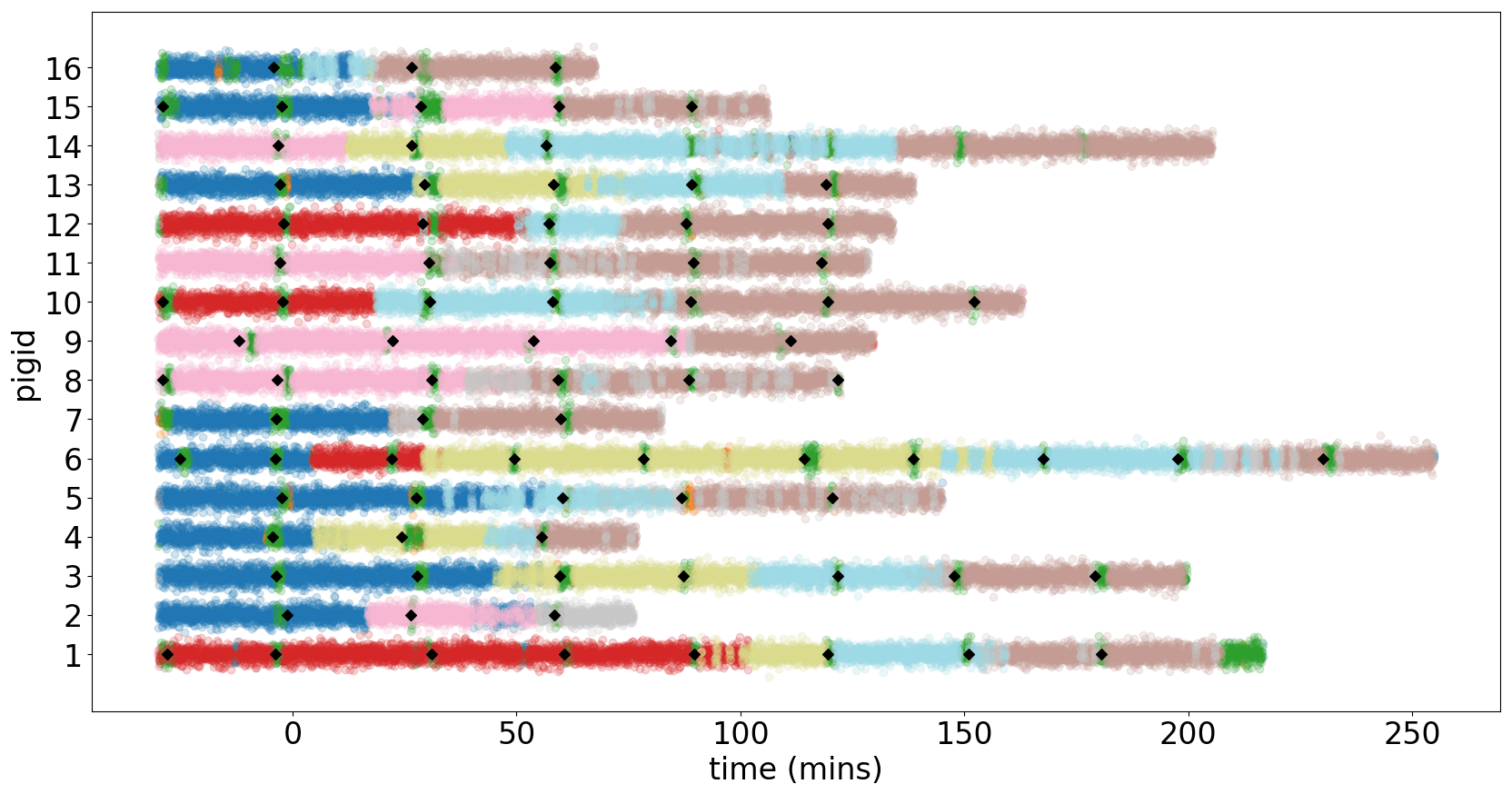}
    \caption{10 clusters, considering intra-subject and inter-subject differences}
    \includegraphics[width=\linewidth]{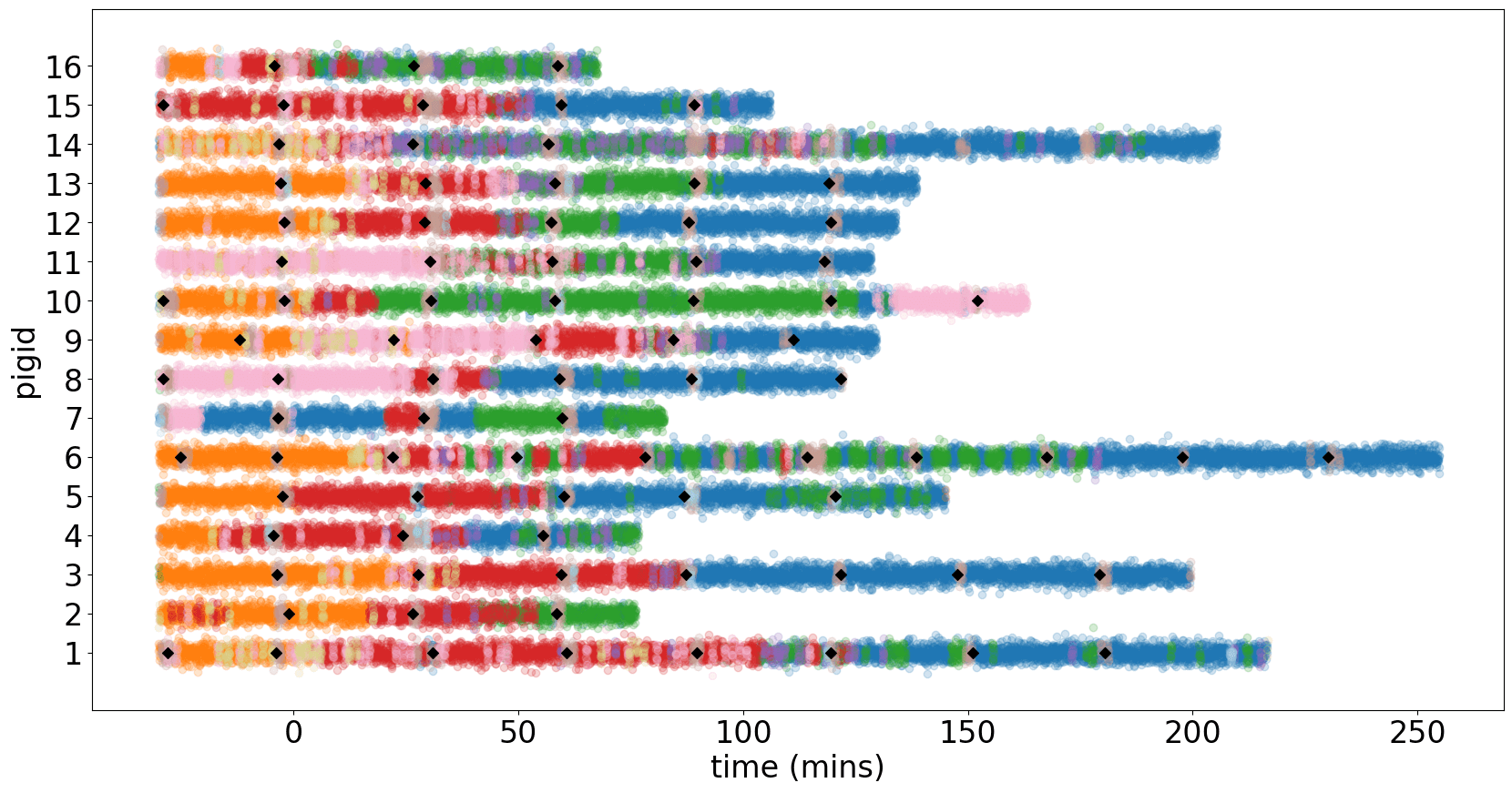}
    \caption{10 clusters, considering intra-subject differences only}
\end{subfigure}
\caption{Resulting clusters mapped through agglomerative clustering with Ward linkage~\cite{liao2005clustering} of 128 dimensional latent embeddings of the vital signs during induced bleed sequences over time. The bleed starts at time $t=0$. Thus, negative time indicates the prebleed, stable resting period. The black dots denote laboratory blood draw events. The colors represent the different clusters found by the algorithm.}
\label{fig:pigsbytime_diff}
\end{figure*}

Fig.~\ref{fig:pigsbytime_diff} shows our latent embedding clusters plotted over time and split by subject. When clustering with 2-3 groups ((a) and (b)), we see a cluster that corresponds to the healthy state (light blue), shifting into another cluster that corresponding to an unhealthy (bleeding) state (brown and dark blue). If the algorithm considers both inter-subject and intra-subject differences (a), using just two clusters is impractical as one cluster is fit entirely to the laboratory blood draws (the clusters near the black dots). This may be due to the model having to learn extra information compared to the model only considering intra-subject differences. One interesting observation is that the transition from the first cluster to the second cluster is not immediate -- the delay in observed physiologic state is presumably, because self-defense response mechanisms triggered by the subject's physiology early in the hemorrhage make this change more subtle to detect. 

When clustering with 10 groups ((c) and (d)), we receive many clusters, yet, they have little overlap in time (compared to 11 or 12 groups). We are also able to discern differences in cluster progression throughout the bleed between different subjects; however, all subjects generally end up in the same eventual state. This follows clinical intuition that the hemodynamic presentations among subjects tend to show substantial heterogeneity when they are stable, but they become more homogeneous the stress escalates.
Additionally, we are able to detect blood draws. They show up as noise (the cluster that corresponds with the black dots) that occurs regularly approximately every 30 minutes in the plot of the latent clusters. 

Another interesting observation is that even when trained to only distinguish differences from a single subject at a time, the model still learns that there are clusters of subjects that are more similar in the initial state than other subjects. We also found that hemodynamic responses between subjects are not universal, as some subjects skip clusters or start their response to bleed in different clusters compared to other subjects. For example, subjects 10 and 12 start off in the same cluster, but 11 is in a different cluster for both sampling methods in Fig~\ref{fig:pigsbytime_diff} (c),(d). We also see that subjects can go through as many as 5 and as low as 2 distinct clusters in the process.

The confusion matrix for the Random Forest Classifier in Fig.~\ref{fig:confusion_matrix} shows that the classifier was best able to predict the blue and orange clusters in Fig.~\ref{fig:pigsbytime_diff} (c),(d) which correspond to the start and end of the vital sign sequences for both training methodologies. However, the classifier was unable to predict the intermediate clusters with high accuracy while the simple neural network was able to. This may suggest that the encoder and the 2 layer network were able to learn more sophisticated non-linear relationship between the features compared to the random forest. Considering both intra-subject and inter-subject differences look better in terms of noise in Fig.~\ref{fig:pigsbytime_diff} (c) and also in mean cluster prediction accuracy. This may be because the model absorbs more useful insights in the vital signs to produce a more separated latent space when allowed to consider the differences between subjects. Additionally, we show that our model is also robust. Fig.~\ref{fig:clusters_cv} shows that the encodings predicted on unseen data are also generally clustered similarly to the training subjects. This shows that the model is learning consistent, potentially meaningful, embeddings.

\section{Conclusion}
We presented a proof-of-concept method of discerning patterns of hemodynamic stress response in the raw vital sign data. We found clusters that generally correspond to the start, intermediate, and end of the bleed. Additionally, we found that the initial clusters usually vary among subjects when they are stable but become more homogeneous as the subjects undergo subsequently escalating stress, which corresponds to clinical intuition. When considering 2 clusters only, we found that the shift between the first and second cluster does not occur immediately after the start of bleed. This makes sense as a significant shift in vital signs (may represent a substantial shift in hemodynamic response) to the slow rate of bleed should not be immediate. Further research is necessary to validate the identified clusters, including more quantitatively or theoretically rigorous evaluation metrics for the clusters we observed.


\bibliographystyle{unsrtnat}
\bibliography{bibliography.bib}

\appendix
\section{Appendix}

\subsection{Additional Methdology}
\label{app:Additional Methdology}

\begin{figure}[ht]
    \centering
    \includegraphics[width=.33\linewidth]{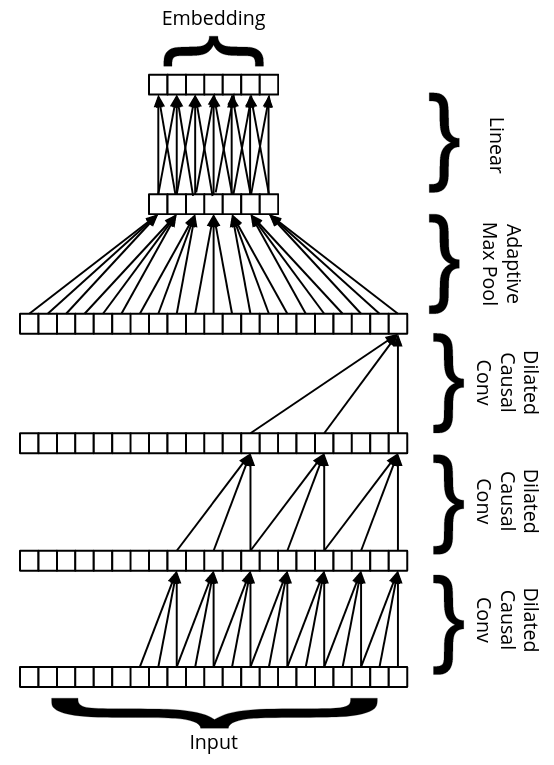}
    \caption{Graphical representation of dilated and causal convolutions. The output of each row is the row above it. The bottom-most row of boxes denotes a variable length input, and the top-most row of boxes an out of latent vector embeddings. While only 3 dilated causal CNN layers shown here, in the actual model, the number of these modules is a hyperparameter that we can specify. The dilation is visualized by the doubling of the gaps between the boxes that is analyzed by the CNN as you go further up in the output layers. The causal cnn is shown by the fact that the final output box on the upper right only has access to the information of the boxes before it--a regular CNN would have access to future timestep information after its current timestep as well. The adaptive pooling layer comes after the dilated causal CNN layers and simply reduces an arbitrary dimensional input to a fixed-sized output. Finally, the final layer is a fully connected layer that outputs the embedding. }
    \label{fig:model}
\end{figure}

The data consists of health metrics of 93 pigs in total. These pigs are separated into 4 groups, which are then bleed at different rates - 60mL/min, 20mL/min, 5mL/min, and 0mL/min respectively. Each pig was monitored for 11 vital signs at 250 Hz (synchronized): arterial and venous blood pressures (CVP, arterial pressure fluid filled and millar, pulmonary pressure), arterial and venous oxygen saturations (SpO2, SvO2), EKG, Plethysmograph, CCO, stroke volume variation (Vigeleo), and airway pressure. The data collection methodology is similar to \citet{pinsky1984instantaneous}'s.

Additionally, we also explored other clustering techniques that we weren't able to show due to page constraints. However, the results we obtained from these methods were not completely different than graphs that we showed previously in Fig.~\ref{fig:pigsbytime_diff}. With the time embeddings, repeating of the clusters over time practically disappeared. We explored:
\begin{enumerate}
    \item Clustering methods (All of these are implemented in sklearn \cite{scikit-learn})
    \begin{enumerate}
        \item K-means
        \item Agglomerative clustering with ward linkage (Bottom-up hierarchical clustering. Ward's linkage merges the two clusters such that the increase in the value of the sum-of-squares variance is minimized 
        \cite{liao2005clustering}. Specifically, sklearn references \cite{mullner2011modern}.)
        \item Gaussian Mixture Models - They performed well and should be analyzed in future research.
        \item Spectral Clustering - We ran into problems with computational limits, so future studies should attempt to resolve this.
    \end{enumerate}
    \item Weighted Time Embeddings (scaled to be less than 2 standard deviations of the embeddings) added to latent embeddings. The type of time embedding is taken from attention transfomers, from \citet{vaswani2017attention}. Note that adding time information allowed training without considering intra-subject, and inter-subject differences to not have repeat clusters over time, but there are no guarantees that the model isn't overfitting to the time embeddings.
    \begin{enumerate}
        \item No time information added
        \item Adding time information for full length of the sequence
        \item Adding time information only from the start of bleed (Since we know the exact location of the start of bleed from the physician annotations, we only add temporal information to the embeddings obtained after the subject starts bleeding. The prebleed embeddings are left alone. This is effectively adding information about the amount of blood lost, since bleed speed is constant after the subject starts bleeding).
    \end{enumerate}
    \item 64, 128, and 256 dimensional latent embeddings. We chose 128 dimensional embeddings since clusters looked the best compared to 64 dimensional--wasn't able to learn more than 3 clusters per subject, and 256 dimensional--too noisy.
    \item Training schemes. These can be implemented through the sampling algorithm \ref{alg:sampling} by passing in various batch sizes.
    \begin{enumerate}
        \item Allowing the encoder to discriminate between subjects (i.e.  allowing sampling negative samples from different subject bleed time series in the batch, this allows model to learn inter-subject differences as well as being able to learn differences the raw data over time).
        \item Disallowing the encoder to discriminate between subjects (negative samples will only be sampled from the same subject, model is restricted to only being able to learn differences in the raw data over time).
    \end{enumerate}
    \item Number of clusters: In addition to all of these methods, we also explore 11 different numbers of clusters that we pass into the clustering algorithms (from 2 to 12 clusters).
\end{enumerate}

\subsection{Triplet Loss}
\label{app:Triplet Loss}

We will use triplet loss to train our encoder as specified by \citet{franceschi2019unsupervised}. Triplet loss is a loss function with a natural intuition as its basis--similar things should embeddings that are closer together than embeddings from unsimilar things. This is reflected in the following loss function. Let $f$ be our encoder that obtains latent vectors from the time series data. Let $x, x^{pos}, x^{neg}_k$ be the reference time series, a positive time series example, and a negative time series example (see  Alg.\ref{alg:sampling} and Fig.~\ref{fig:sampling}). Let K be the number of negative samples to take. Then, the loss is shown in Eqn.~\ref{eqn:triplet}.
\begin{equation}
\label{eqn:triplet}
 L=-log(\sigma(f(x)^Tf(x^{pos})))-\sum^K_{k=1}log(\sigma(-f(x)^Tf(x^{neg}_k)))
\end{equation}
Triplet loss is popular in natural language processing - Word2vec \cite{mikolov2013distributed} as it is effective in training unsupervised models that obtain latent vectors from words that encode some semantic meaning. \citet{franceschi2019unsupervised}. demonstrated that this is useful for unsupervised learning of useful embeddings of general multivariate time sequences as well.

\subsection{Sampling methodology}
\label{app:Sampling methodology}

\begin{figure}[h!]
    \centering
    \includegraphics[width=.5\linewidth]{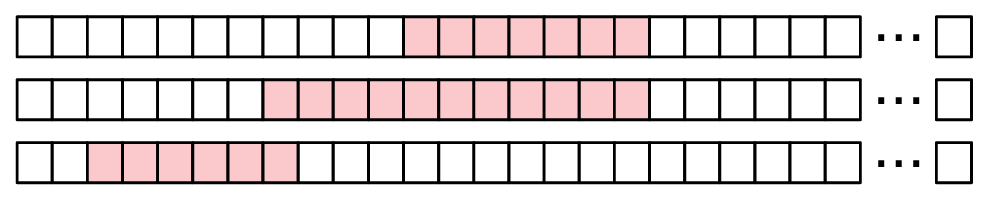}
    \caption{Sampling methodology. Let the each row of boxes represent multivariate sequences of our training data, and let the rows be the sequences in the batch where we choose the references from. Then, the pink squares represent the reference timesteps, positive timesteps are sampled form within the reference timesteps. Negative timesteps are sample from the white boxes (all of the timesteps excluding the reference timesteps).}
    \label{fig:sampling}
\end{figure}

We use a modified version of \citet{franceschi2019unsupervised}s sampling algorithm to obtain choices of reference $x$, positive example $x^{pos}$, and negative example $x^{neg}_k$. This is different from the original implementation from \citet{franceschi2019unsupervised}. since the negative samples are only choosen randomly when there can be no overlap with the reference time series; thus, this guarantees that a negative example can't be a positive example as well. This should allow the model to learn better as there is a clearer difference between positive and negative samples. Algorithm \ref{alg:sampling} and Fig.~\ref{fig:sampling} shows proposed sampling algorithm for one gradient update. Once our samples have been taken, we pass them into our dilated causal cnn to obtain embeddings and then update the weights of the network using the loss function in Equation \ref{eqn:triplet}.

\begin{algorithm}[h!]
\caption{\\Input: A training batch of sequences $y_i$, where $i$ is the ith training sequence. Let N be the total number of sequences in this training batch. Let K be the number of negative samples to be sampled per batch item.\\\\
Output: N reference samples $x^{ref}$, N samples of positive samples $x^{pos}$, and N*K negative samples $x^{neg}$.} 
\label{alg:sampling}
\begin{algorithmic}
\FOR{$i \in [1,N]$}
    \STATE randomly choose length of reference sample $len_i^{ref} \in [1, len(y_i)]$
    \STATE randomly choose length of positive sample $len_i^{pos} \in [1, len_i^{ref}]$
    \STATE randomly choose reference sample $x_i^{ref}$ from subseries of $y_i$ of length $len_i^{ref}$
    \STATE randomly choose positive sample $x_i^{pos}$ from subseries of $x_i^{ref}$ of length $len_i^{pos}$
\ENDFOR

\FOR {$k \in [1, K*N]$}
    \STATE randomly choose $y_k$ from the batch
    \STATE randomly choose $len^{neg}_k \in [1, len(y_k)]$
    \STATE Let $x_{y_{k}}^{ref}$ be the reference sample that we previously took from $y_k$
    \STATE randomly choose $x^{neg}_k$ among subseries of $y_{k}$ of length $len^{neg}_k$ s.t. $x^{neg}_k \cap x_{y_{k}}^{ref} = \emptyset$
\ENDFOR
\end{algorithmic}
\end{algorithm}

\subsection{Additional Results}
\begin{figure*}[h]
\centering
\begin{subfigure}[h]{0.49\textwidth}
    \includegraphics[width=\linewidth]{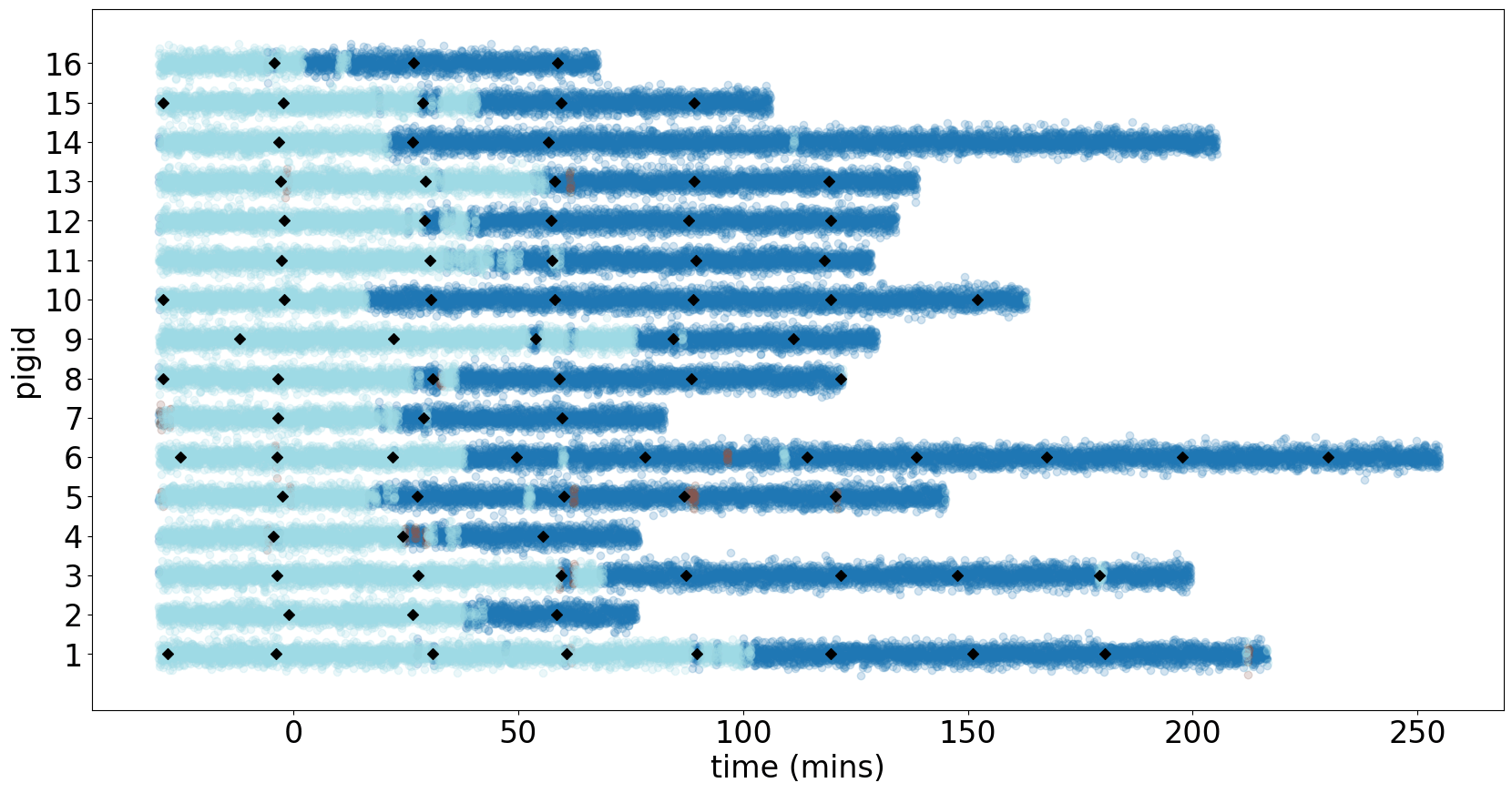}
    \caption{3 clusters, considering intra-subject and inter-subject differences}
    \includegraphics[width=\linewidth]{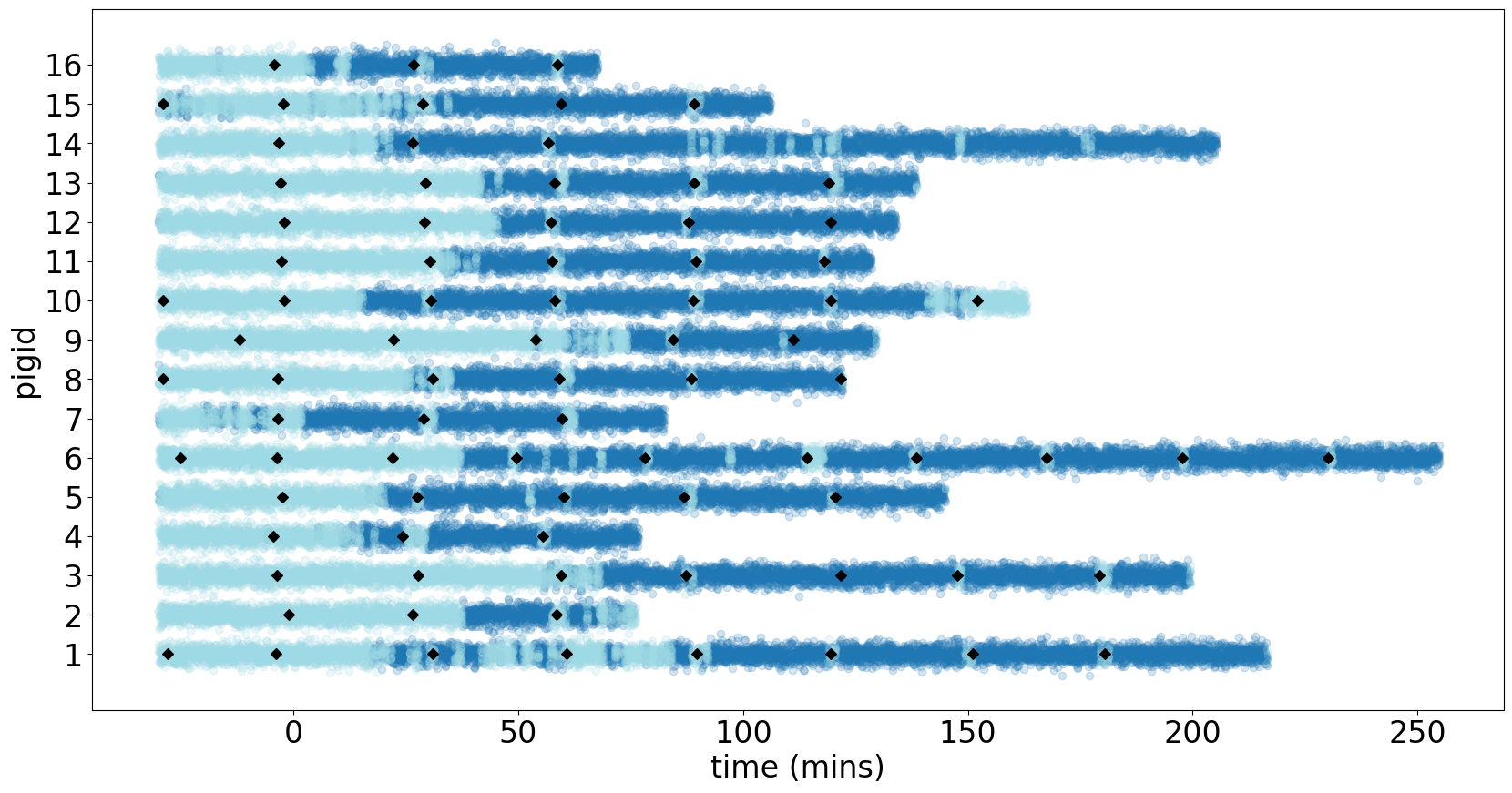}
    \caption{2 clusters, considering intra-subject differences only}
\end{subfigure}%
\hfill
\begin{subfigure}[h]{0.49\textwidth}
    \includegraphics[width=\linewidth]{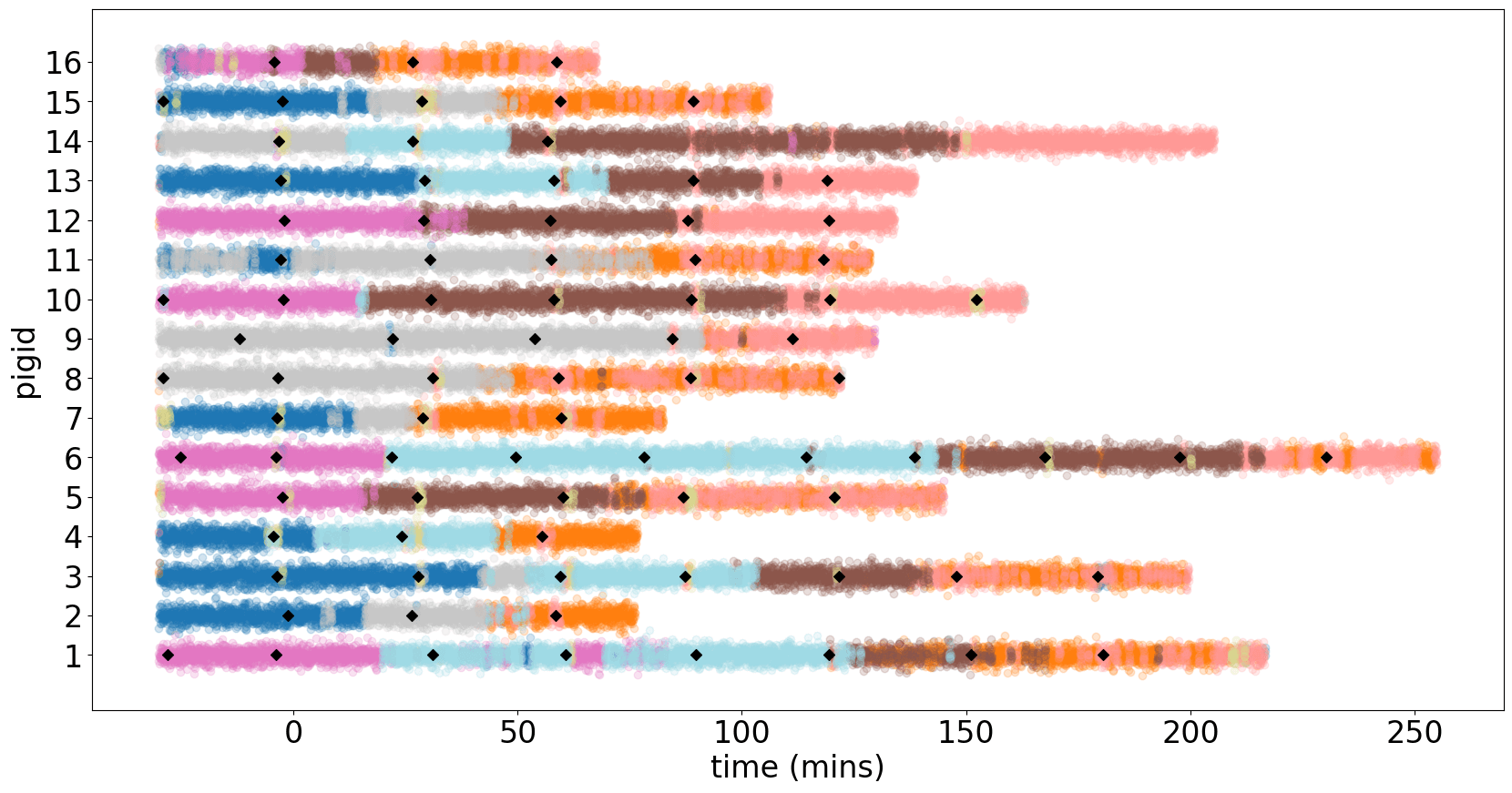}
    \caption{10 clusters, considering intra-subject and inter-subject differences}
    \includegraphics[width=\linewidth]{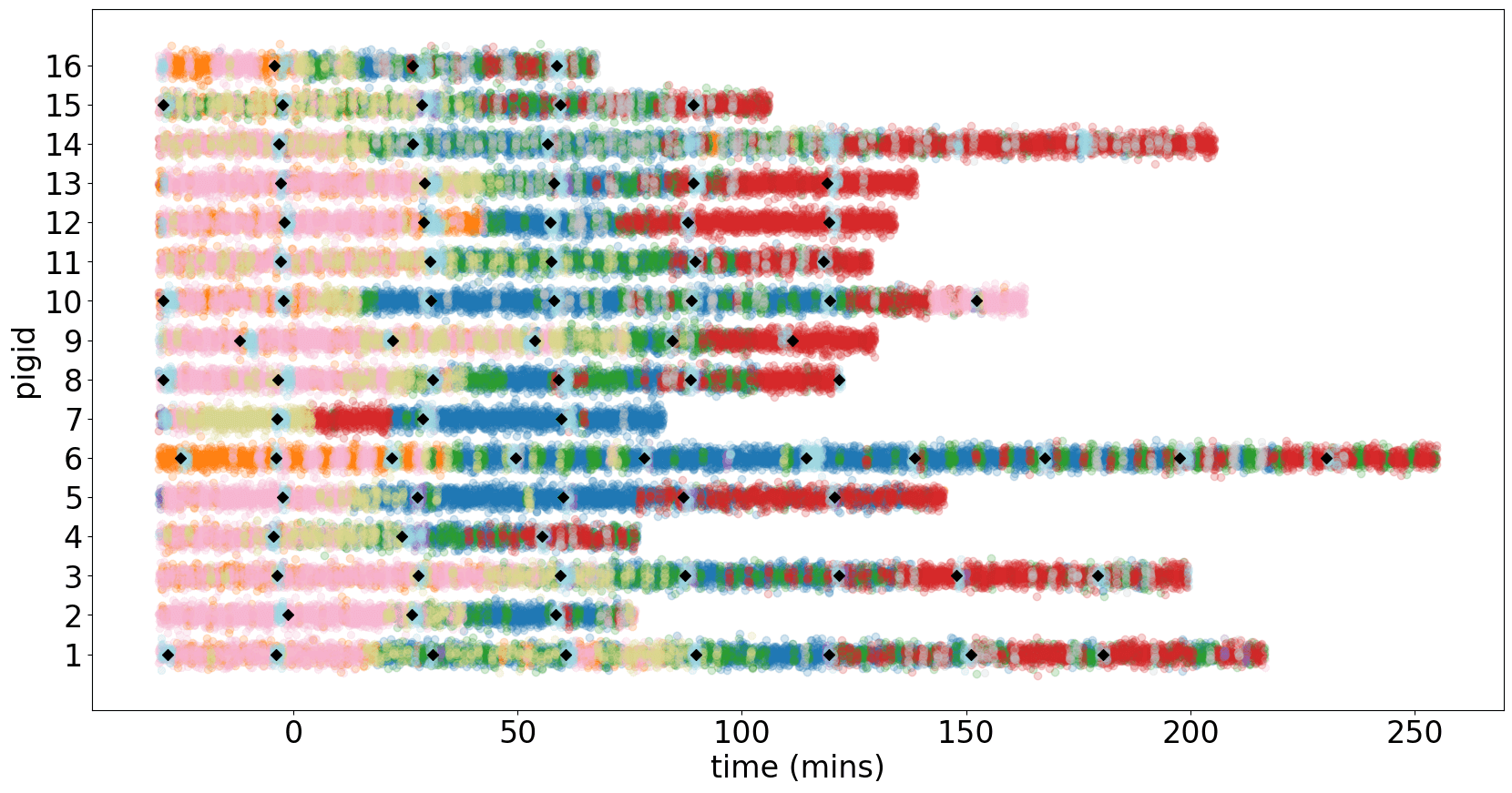}
    \caption{10 clusters, considering intra-subject differences only}
\end{subfigure}
\caption{The plots of the subjects' clusters (through $K$-means clustering) of 128 dimensional latent embeddings of the bleed sequence over time. The bleed starts at time=0, so negative time indicates the prebleed "stable" period. The black dots are the times where noisy blood draws occur. The colors represent the different clusters found by the clustering algorithm. In general, the $K$-means cluster patterns over time look similar to the agglomerative cluster patterns.}
\label{fig:clusters_cv}
\end{figure*}

\begin{figure*}[h]
\centering
\begin{subfigure}[h]{0.49\textwidth}
    \includegraphics[width=\linewidth]{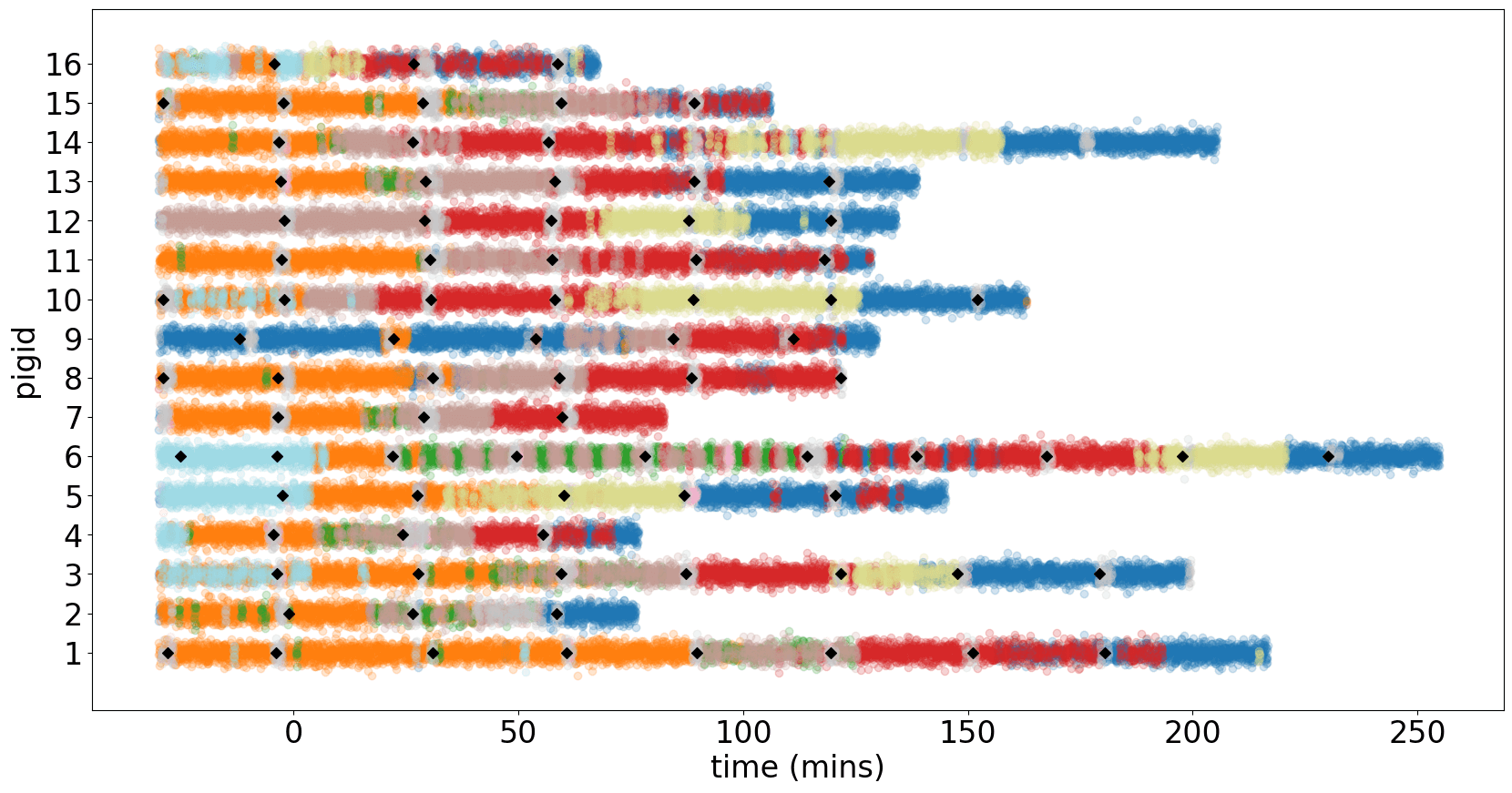}
    \caption{Fold 1, trained on subjects 1-12}
    \includegraphics[width=\linewidth]{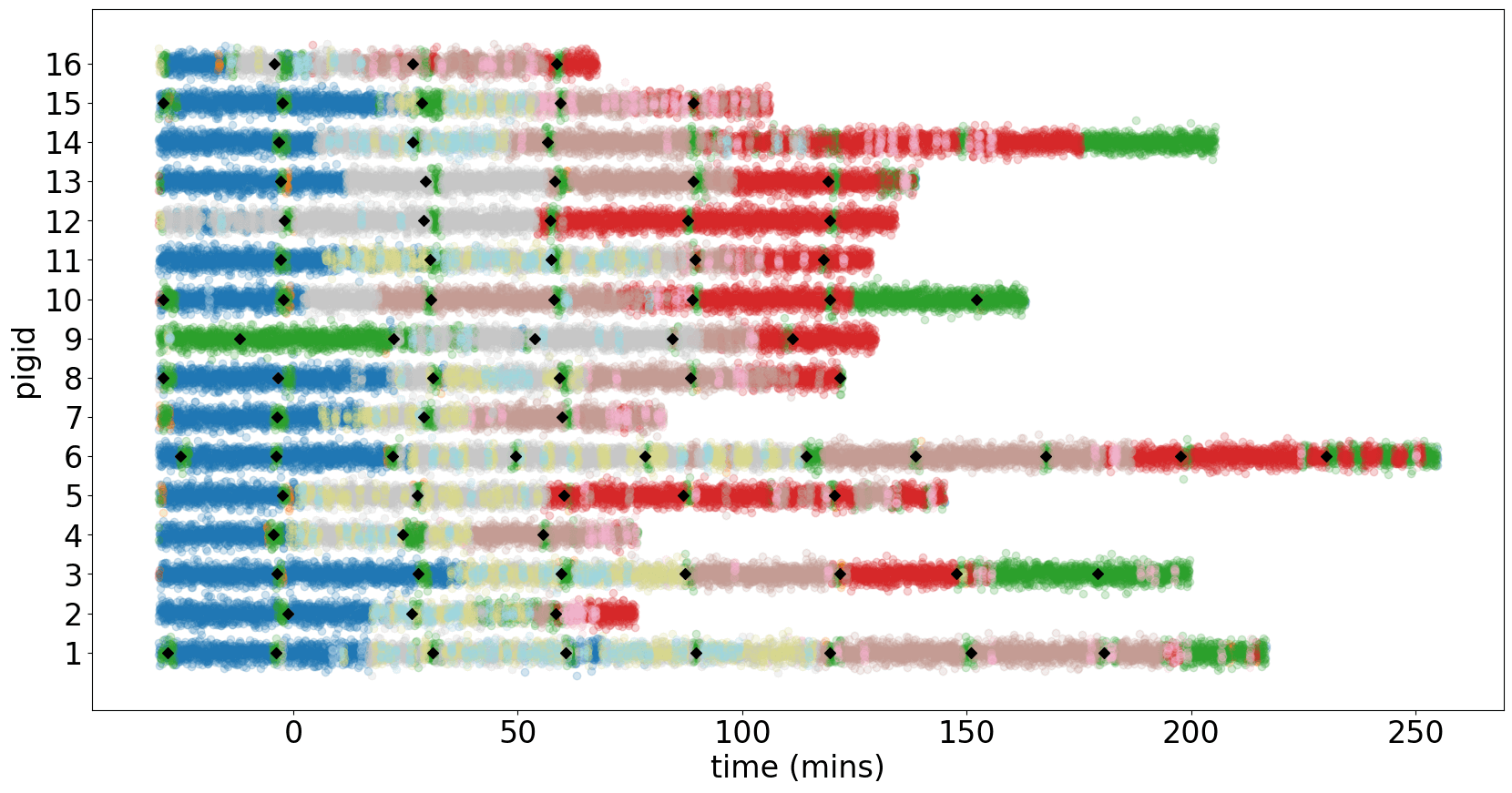}
    \caption{Fold 2, trained on subjects 1-8 and 13-16}
\end{subfigure}%
\hfill
\begin{subfigure}[h]{0.49\textwidth}
    \includegraphics[width=\linewidth]{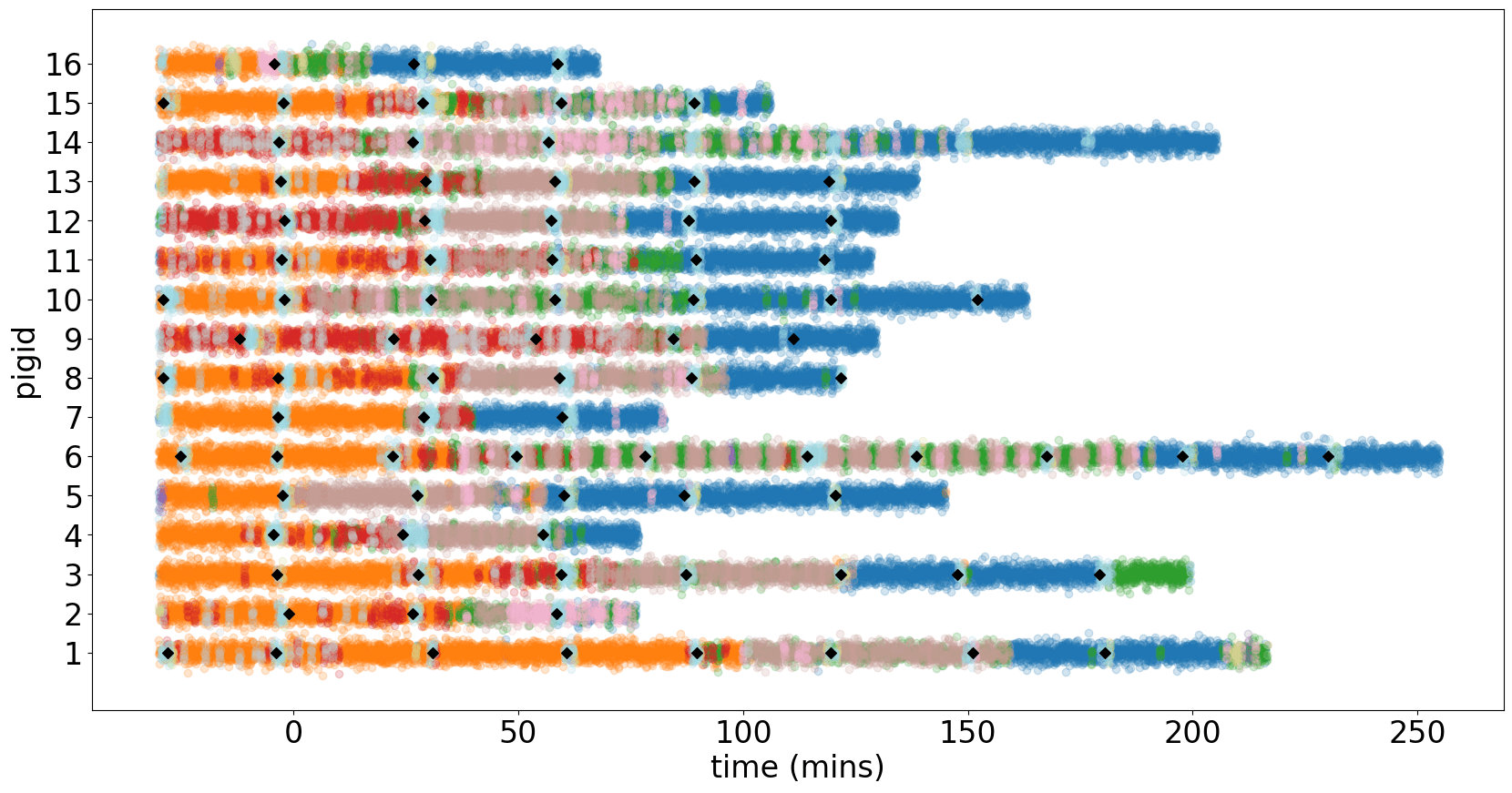}
    \caption{Fold 3, trained on subjects 1-4 and 8-16}
    \includegraphics[width=\linewidth]{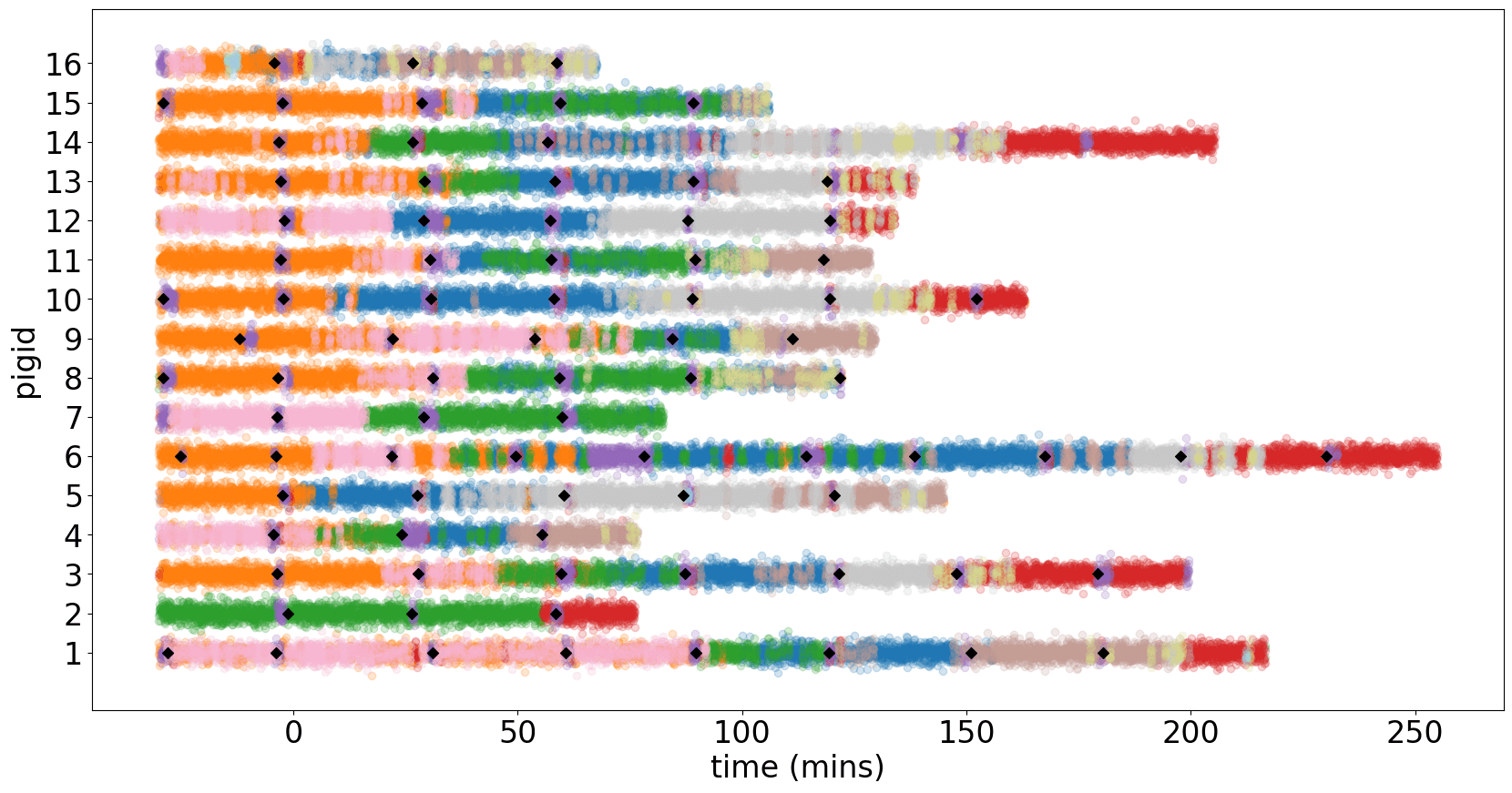}
    \caption{Fold 4, trained on subjects 5-16}
\end{subfigure}
\caption{The plots of the subjects' clusters of 128 dimensional latent embeddings of the bleed sequence over time with 4-fold train/test split. After training, we used the encoder to obtain latent embeddings of all of the subjects and clustered them via agglomerative clustering (10 clusters). The bleed starts at time $t=0$, so negative time indicates the prebleed "stable" period. The black dots are the times where noisy blood draws occur. The colors represent the different clusters found by the clustering algorithm. For the most part, the predicted clusters for the test subjects plotted over had a similar order as the training subjects.}
\label{fig:pigsbytime_diff_kmeans}
\end{figure*}

\begin{figure*}[h]
    \centering
    \begin{subfigure}[h]{.75\textwidth}
        \includegraphics[width=\linewidth]{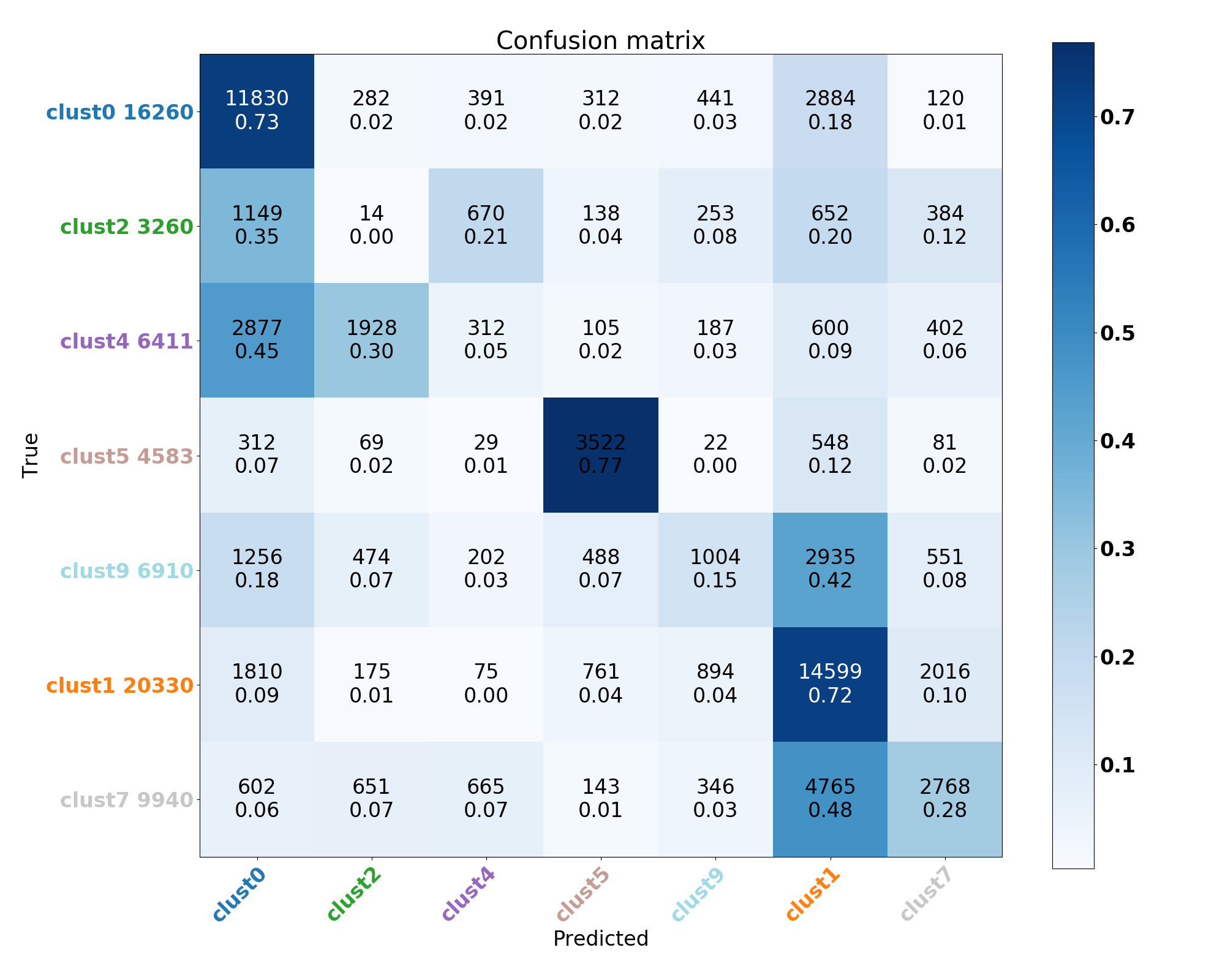}
        \caption{Random Forest}
    \end{subfigure}%
    \hfill
    \begin{subfigure}[h]{.75\textwidth}
        \includegraphics[width=\linewidth]{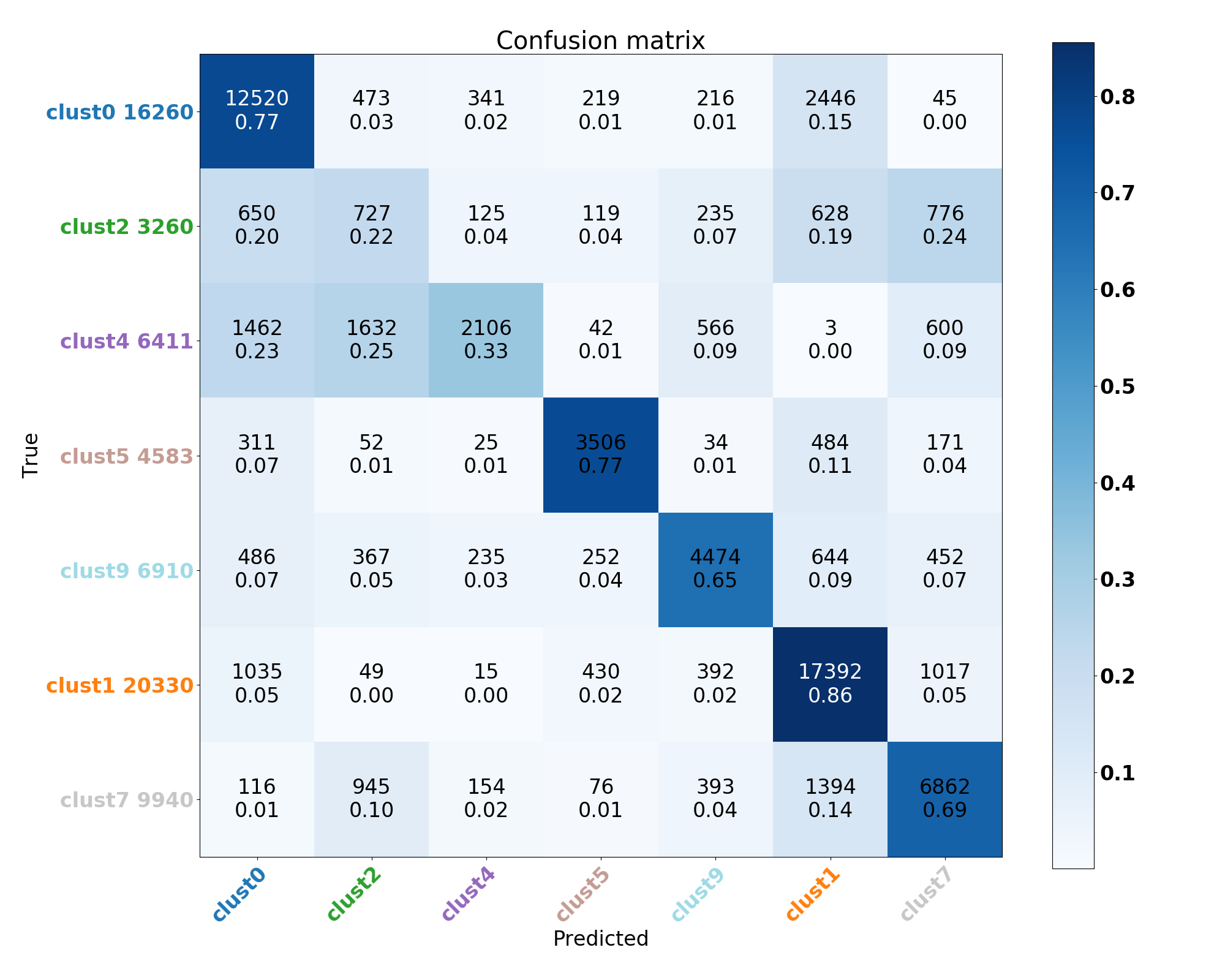}
        \caption{Neural Network}
    \end{subfigure}
    \caption{Aggregate Confusion Matrix of a Random Forest Classifier trained on 16 fold (by subject) cross validation to predict agglomerative clustering labels for 10 clusters. The training data consists of means, medians, standard deviations, 2.5 percentile, 97.5 percentile, the 95 percentile range, and the binned (10 bins) max power over a power-frequency transform of all of the variables. This considers intra-pig and inter-pig differences.}
    \label{fig:confusion_matrix}
\end{figure*}

See Fig.~\ref{fig:pigsbytime_diff_kmeans} for a clustering with K-means. Fig.~\ref{fig:clusters_cv} shows a 4-fold cross-validation to show the robustness of our method, Table~\ref{tab:cluster_classification} shows and example of the accuracies that we use for the confusion matrix.  Fig.~\ref{fig:confusion_matrix} shows the confusion matrix for our Random Forest and 2-layer Neural Network models.

\begin{table}[h]
\begin{subtable}[h]{\textwidth}
\resizebox{\linewidth}{!}{
\begin{tabular}{@{}ccccccccccccccccccc@{}}
clusters & pig1 & pig2 & pig3 & pig4 & pig5 & pig6 & pig7 & pig8 & pig9 & pig10 & pig11 & pig12 & pig13 & pig14 & pig15 & pig16 & mean & std \\ \midrule
12 & 0.19 & 0.29 & 0.45 & 0.48 & 0.32 & 0.42 & 0.52 & 0.34 & 0.26 & 0.73 & 0.74 & 0.48 & 0.59 & 0.40 & 0.66 & 0.52 & 0.46 & 0.16 \\
11 & 0.24 & 0.48 & 0.55 & 0.63 & 0.46 & 0.31 & 0.57 & 0.39 & 0.24 & 0.76 & 0.76 & 0.55 & 0.59 & 0.39 & 0.67 & 0.53 & 0.51 & 0.16 \\
10 & 0.23 & 0.58 & 0.60 & 0.63 & 0.45 & 0.38 & 0.80 & 0.62 & 0.32 & 0.80 & 0.73 & 0.49 & 0.70 & 0.54 & 0.79 & 0.54 & 0.57 & 0.16 \\
9 & 0.56 & 0.57 & 0.58 & 0.63 & 0.64 & 0.41 & 0.78 & 0.58 & 0.33 & 0.81 & 0.72 & 0.56 & 0.67 & 0.45 & 0.78 & 0.55 & 0.60 & 0.13 \\
8 & 0.62 & 0.58 & 0.57 & 0.62 & 0.63 & 0.41 & 0.78 & 0.52 & 0.31 & 0.78 & 0.74 & 0.51 & 0.62 & 0.47 & 0.79 & 0.54 & 0.59 & 0.13 \\
7 & 0.55 & 0.53 & 0.64 & 0.62 & 0.49 & 0.57 & 0.81 & 0.44 & 0.25 & 0.79 & 0.75 & 0.41 & 0.70 & 0.61 & 0.70 & 0.54 & 0.59 & 0.14 \\
6 & 0.66 & 0.66 & 0.63 & 0.61 & 0.50 & 0.66 & 0.79 & 0.55 & 0.33 & 0.80 & 0.75 & 0.63 & 0.67 & 0.49 & 0.77 & 0.90 & 0.65 & 0.13 \\
5 & 0.68 & 0.60 & 0.65 & 0.66 & 0.48 & 0.75 & 0.82 & 0.46 & 0.29 & 0.90 & 0.71 & 0.65 & 0.69 & 0.53 & 0.75 & 0.88 & 0.66 & 0.16 \\
4 & 0.94 & 0.53 & 0.93 & 0.95 & 0.66 & 0.96 & 0.93 & 0.93 & 0.35 & 0.95 & 0.79 & 0.77 & 0.96 & 0.76 & 0.88 & 0.93 & 0.83 & 0.17 \\
3 & 0.93 & 0.58 & 0.95 & 0.92 & 0.65 & 0.96 & 0.93 & 0.92 & 0.35 & 0.94 & 0.79 & 0.78 & 0.94 & 0.74 & 0.89 & 0.93 & 0.83 & 0.17 \\
2 & 1.00 & 1.00 & 0.99 & 0.98 & 0.99 & 1.00 & 0.99 & 0.99 & 1.00 & 1.00 & 1.00 & 1.00 & 0.99 & 1.00 & 1.00 & 1.00 & 1.00 & 0.01 \\ \hline
\end{tabular}}
\caption{Considering intra-pig and inter-pig differences}
\end{subtable}
\begin{subtable}[h]{\textwidth}
\resizebox{\linewidth}{!}{
\begin{tabular}{@{}ccccccccccccccccccc@{}}
clusters & pig1 & pig2 & pig3 & pig4 & pig5 & pig6 & pig7 & pig8 & pig9 & pig10 & pig11 & pig12 & pig13 & pig14 & pig15 & pig16 & mean & std \\ \hline
12 & 0.56 & 0.29 & 0.75 & 0.43 & 0.60 & 0.41 & 0.17 & 0.30 & 0.51 & 0.27 & 0.43 & 0.63 & 0.60 & 0.36 & 0.33 & 0.22 & 0.43 & 0.16 \\
11 & 0.54 & 0.49 & 0.73 & 0.42 & 0.42 & 0.57 & 0.40 & 0.32 & 0.37 & 0.36 & 0.38 & 0.60 & 0.61 & 0.34 & 0.22 & 0.45 & 0.45 & 0.13 \\
10 & 0.68 & 0.35 & 0.86 & 0.60 & 0.57 & 0.59 & 0.25 & 0.40 & 0.41 & 0.28 & 0.38 & 0.69 & 0.61 & 0.41 & 0.38 & 0.38 & 0.49 & 0.16 \\
9 & 0.55 & 0.39 & 0.78 & 0.50 & 0.54 & 0.59 & 0.26 & 0.39 & 0.42 & 0.36 & 0.44 & 0.64 & 0.58 & 0.40 & 0.36 & 0.37 & 0.47 & 0.13 \\
8 & 0.46 & 0.62 & 0.74 & 0.43 & 0.50 & 0.46 & 0.38 & 0.39 & 0.34 & 0.48 & 0.49 & 0.65 & 0.63 & 0.41 & 0.26 & 0.45 & 0.48 & 0.12 \\
7 & 0.65 & 0.75 & 0.78 & 0.59 & 0.58 & 0.66 & 0.45 & 0.38 & 0.46 & 0.59 & 0.61 & 0.73 & 0.73 & 0.51 & 0.30 & 0.46 & 0.58 & 0.14 \\
6 & 0.67 & 0.88 & 0.75 & 0.59 & 0.61 & 0.59 & 0.47 & 0.58 & 0.64 & 0.69 & 0.67 & 0.72 & 0.78 & 0.49 & 0.42 & 0.65 & 0.64 & 0.11 \\
5 & 0.65 & 0.85 & 0.76 & 0.62 & 0.64 & 0.62 & 0.50 & 0.61 & 0.55 & 0.73 & 0.69 & 0.72 & 0.81 & 0.47 & 0.42 & 0.65 & 0.64 & 0.11 \\
4 & 0.73 & 0.52 & 0.85 & 0.71 & 0.63 & 0.81 & 0.44 & 0.87 & 0.42 & 0.69 & 0.70 & 0.79 & 0.79 & 0.60 & 0.43 & 0.65 & 0.66 & 0.14 \\
3 & 0.76 & 0.63 & 0.83 & 0.76 & 0.68 & 0.82 & 0.43 & 0.87 & 0.80 & 0.65 & 0.76 & 0.84 & 0.84 & 0.53 & 0.61 & 0.78 & 0.72 & 0.12 \\
2 & 0.79 & 0.84 & 0.94 & 0.89 & 0.88 & 0.94 & 0.63 & 0.94 & 0.51 & 0.81 & 0.88 & 0.85 & 0.85 & 0.84 & 0.56 & 0.76 & 0.81 & 0.13 \\ \hline
\end{tabular}}
\caption{Considering intra-subject differences only}
\end{subtable}
\caption{Accuracy of a Random Forest Classifier trained on 16 fold (by subject) cross validation to predict agglomerative clustering labels. The training data consists of means, medians, standard deviations, 2.5 percentile, 97.5 percentile, the 95 percentile range, and the binned (10 bins) max power over a power-frequency transform of all of the variables.}
\label{tab:cluster_classification}
\end{table}

\subsection{Future work}

In future work, we aim at exploring other encoder architectures, such as Variational Autoencoders \cite{fabius2014variational}, Denoising Autoencoders \cite{vincent2008extracting}, BERT \cite{devlin2018bert}. We can also find clusters or separations using Hidden Markov Models (HMMs) or change point detection methods known from multivariate forecasting. Layer-wise Relevance Propagation \cite{samek2017explainable}, a method that enables the model to explain the output in terms of its input, could allow us to interpret the latent embeddings in terms of the raw input time series. Finally, our goal is to have physicians carefully analyze the validity of the different clusters that we find. 

\end{document}